\documentclass{article}

\PassOptionsToPackage{numbers,compress}{natbib}
\usepackage[preprint]{neurips_2026}

\usepackage[utf8]{inputenc}
\usepackage[T1]{fontenc}
\usepackage[hyphens]{url}
\usepackage{hyperref}
\usepackage{booktabs}
\usepackage{amsfonts}
\usepackage{nicefrac}
\usepackage{microtype}
\usepackage{xcolor}
\usepackage{graphicx}
\usepackage{enumitem}
\usepackage{array}
\usepackage{tikz}
\usetikzlibrary{positioning,arrows.meta,shapes.geometric}
\usepackage[most]{tcolorbox}
\usepackage[normalem]{ulem}

\definecolor{capgreen}{HTML}{000000}
\definecolor{limred}{HTML}{000000}
\definecolor{overcolor}{RGB}{67,75,97}
\definecolor{undercolor}{RGB}{67,75,97}

\newtcolorbox{overestbox}{
  enhanced, colback=overcolor!4, colframe=overcolor!4, boxrule=0pt,
  left=8pt, right=8pt, top=4pt, bottom=4pt, arc=2pt,
  before skip=6pt, after skip=6pt,
  before upper={\textbf{$\uparrow$\, Why benchmarks can \emph{overestimate} capabilities.}\par\smallskip},
}
\newtcolorbox{underestbox}{
  enhanced, colback=undercolor!4, colframe=undercolor!4, boxrule=0pt,
  left=8pt, right=8pt, top=4pt, bottom=4pt, arc=2pt,
  before skip=6pt, after skip=6pt,
  before upper={\textbf{$\downarrow$\, Why benchmarks can \emph{underestimate} capabilities.}\par\smallskip},
}

\newtcolorbox{taxbox}{
  enhanced, breakable, colback=white, colframe=overcolor!40, boxrule=1pt,
  left=8pt, right=8pt, top=4pt, bottom=4pt, arc=2pt,
  before skip=8pt, after skip=8pt,
  before upper={\textbf{Five dimensions for classifying capability evaluations}\par\smallskip},
}

\newcounter{recommendation}
\newtcolorbox{claimbox}[1][]{
    enhanced,
    colback=white,
    colframe=white,
    boxrule=0pt,
    borderline west={2.5pt}{0pt}{black!70},
    left=8pt,
    right=0pt,
    top=0pt,
    bottom=0pt,
    sharp corners,
    before skip=4pt,
    after skip=5pt,
    before upper={\refstepcounter{recommendation}\textbf{Recommendation \therecommendation\ (#1).}\space\itshape},
}

\definecolor{chip1}{HTML}{BFDBFE}
\definecolor{chip2}{HTML}{FEF08A}
\definecolor{chip3}{HTML}{BBF7D0}
\newcommand{\phasechip}[1]{{\setlength{\fboxsep}{0pt}\fcolorbox{black!40}{#1}{\rule{0pt}{0.7em}\rule{0.7em}{0pt}}}}

\title{Open-World Evaluations\\ for Measuring Frontier AI Capabilities}

\author{%
\textbf{Sayash Kapoor}$^{1}$\thanks{Correspondence to: \texttt{\{sayashk, pk7019, rabanser, arvindn\}@princeton.edu}} \quad \textbf{Peter Kirgis}$^{1}$ \quad \textbf{Andrew Schwartz}$^{1,2}$ \quad \textbf{Stephan Rabanser}$^{1}$ \\
\textbf{J.J.~Allaire}$^{3}$ \quad \textbf{Rishi Bommasani}$^{4}$ \quad \textbf{Harry Coppock}$^{5}$ \quad \textbf{Magda Dubois}$^{5}$ \quad \textbf{Gillian Hadfield}$^{6}$ \\
\textbf{Andy Hall}$^{4}$ \quad \textbf{Sara Hooker}$^{7}$ \quad \textbf{Seth Lazar}$^{6,8}$ \quad \textbf{Steve Newman}$^{9}$ \\
\textbf{Dimitris Papailiopoulos}$^{10,11}$ \quad \textbf{Shoshannah Tekofsky}$^{12}$ \quad \textbf{Helen Toner}$^{13}$ \quad \textbf{Cozmin Ududec}$^{5}$ \\
\textbf{Arvind Narayanan}$^{1}$ \\
{\normalfont\small $^{1}$Princeton University \quad $^{2}$Cornflower Labs \quad $^{3}$Meridian Labs \quad $^{4}$Stanford University} \\
{\normalfont\small $^{5}$UK AI Security Institute \quad $^{6}$Johns Hopkins University \quad $^{7}$Adaption Labs} \\
{\normalfont\small $^{8}$Australian National University \quad $^{9}$Golden Gate Institute for AI \quad $^{10}$UW Madison} \\
{\normalfont\small $^{11}$Microsoft Research \quad $^{12}$AI Digest \quad $^{13}$Georgetown University (CSET)}
}

\begin{document}
\renewcommand{\thefootnote}{\fnsymbol{footnote}}
\maketitle
\setcounter{footnote}{0}
\renewcommand{\thefootnote}{\arabic{footnote}}

\begin{abstract}
Benchmark-based evaluation remains important for tracking frontier AI progress. But it can both overstate and understate deployed capability because it privileges tasks that can be precisely specified, automatically graded, easy to optimize for, and run with low budgets and short time horizons. We advocate for a complementary class of evaluations, which we term \emph{open-world evaluations}: long-horizon, messy, real-world tasks assessed through small-sample qualitative analysis rather than benchmark-scale automation. In this paper we survey recent open-world evaluations, identify their strengths and limitations, and introduce CRUX (\textbf{C}ollaborative \textbf{R}esearch for \textbf{U}pdating AI e\textbf{X}pectations), a project for conducting such evaluations regularly. As a first instance, we task an AI agent with developing and publishing a simple iOS application to the Apple App Store. The agent completed the task with only a single avoidable manual intervention, suggesting that open-world evaluations can provide early warning of capabilities that may soon become widespread. We conclude with recommendations for designing and reporting open-world evals.
\end{abstract}

\section{Introduction}

Tracking and predicting the capabilities of frontier AI systems is an open methodological problem, with stakes that rise as these systems enter increasingly consequential settings. Currently, the predominant approach relies on benchmarking, where agents are scored on large suites of automatically graded tasks. These benchmark-based evaluations underpin much of the public discussion surrounding AI progress. For instance, METR's time horizon graph \citep{metrOrgTimeHorizons} has been widely cited by industry leaders, safety organizations, and policy analysts as evidence of rapid capability growth. As decisions regarding funding, regulation, and safety investments are increasingly based on these measurements, their accuracy becomes critical, and any systematic gaps in what they capture deserve scrutiny.

Despite this reliance, benchmarks can simultaneously overestimate and underestimate actual progress. Overestimation frequently occurs because any task specified precisely enough to benchmark is also specified precisely enough to optimize for. This dynamic allows agents to excel on a test without necessarily acquiring the underlying capability. Compounding this issue, test sets from prominent benchmarks often leak into training data, occurring either directly or through close paraphrases of held-out tasks. Conversely, underestimation arises when low benchmark accuracy reflects incidental failures rather than genuine capability gaps. An agent fundamentally capable of completing a task might still fail because it encounters a CAPTCHA, hits a rate limit, or gets stuck on a brittle GUI element. Taken together, these issues demonstrate that \emph{benchmark scores conflate the target capability with artifacts of the evaluation environment}. The resulting signal is noisy for the questions decision-makers actually care about, and grows noisier as agents become more capable.

To address these limitations, a growing body of work has begun evaluating agents on long, complex, real-world tasks that extend beyond traditional benchmarks. For example, \citet{anthropicComEngineeringBuilding} used Claude agents to build a C compiler capable of compiling the Linux kernel, a challenge requiring weeks of agent time and substantial scaffolding. In a separate experiment, Anthropic and Andon Labs tasked Claude with managing a small office shop \citep{anthropicComResearchProject2}, exposing the agent to real customers, live inventory, and actual money. Other researchers have run open-ended projects, coordinated multi-agent workflows over several days, or reimplemented substantial pieces of production software (see Section~\ref{an-incomplete-survey-of-open-world-evaluations} for more examples). Despite their variety, these experiments share a common structure: they rely on small sample sizes, permit human intervention when the agent hits an obstacle unrelated to the tested capability, and prioritize qualitative assessment of agent logs over a single aggregate metric.

Due to this unconventional structure, such evaluations are easily dismissed as unscientific. Each typically features a sample size of one, lacks standardization, and cannot be cleanly independently reproduced. These limitations are real, and we do not claim they can be fully overcome. Nevertheless, this emerging class of evaluations, which we refer to as \textbf{open-world evaluations}, provide critical evidence about AI capabilities that standard benchmarks miss. First, they surface early warnings about emerging capabilities, giving institutions and policymakers lead time to build societal resilience and inform strategic decisions about deployment, regulation, and investment. Second, they reveal blind spots in existing benchmarks where automated grading misses the practical substance of a task.

In this paper we conceptualize open-world evaluations, survey prior examples for best practices and pitfalls, and introduce CRUX, a project to run them regularly. Our key contributions and findings are:

\begin{itemize}[leftmargin=9pt, topsep=-3pt, itemsep=3pt, parsep=0pt]
\item
 \textbf{Open-world evaluations are an important emerging class of AI evaluation} (Section~\ref{open-world-evaluations-are-an-important-emerging-class-of-ai-evaluation}){\textbf{.}} As AI systems grow more capable, evaluations designed to elicit frontier capabilities must grow in complexity. Open-world evaluations are the latest stage in this progression.
\item
 \textbf{We introduce CRUX to conduct open-world evaluations systematically, debuting with an end-to-end iOS app deployment experiment} (Sections~\ref{introducing-crux-collaborative-research-for-updating-ai-expectations} and \ref{crux-1-can-ai-agents-autonomously-develop-and-publish-an-ios-app}){\textbf{.}}
 CRUX (Collaborative Research for Updating AI eXpectations) is our effort to conduct rigorous open-world evaluations and operationalize best practices missing from most prior work. Our first experiment tasked an agent with developing and publishing a simple iOS application to the App Store. While many existing benchmarks test isolated coding skills, our interest lay in whether an agent could handle real-world deployment end-to-end—requiring it to navigate complex, unstandardized steps like signing the app, releasing a privacy policy, completing Apple's forms, and shepherding the app through review.
\item
 \textbf{Our main finding: The agent succeeded after one avoidable manual intervention} (Section~\ref{the-final-evaluation}) --- at one point, it forgot where certain credentials were stored. Log analysis yielded a rich set of additional observations, such as the fact that the agent fabricated a fictional phone number for the App Store review process. Of the \$1{,}000 total cost, 97.5\% went to polling for review status, while development itself cost only \$25 in tokens. The app is now live on the App Store. We disclosed our results to Apple four weeks before initial public disclosure of our findings. App store operators should prepare to handle spam submissions, since agents may soon submit applications at scale.
\item
 \textbf{Methodological recommendations for open-world evaluations} (Section~\ref{lessons-and-future-plans}){\textbf{.}} Drawing on CRUX \#1 and a survey of other open-world evaluations, we identify six recommendations: specifying the measurement construct, documenting interventions, analyzing and releasing logs, real-time monitoring, conducting dry runs, and cost reporting. If these principles were adopted as shared norms, open-world evaluations would yield more actionable insights and be easier to build on.
\end{itemize}

\section{The case for open-world evaluations}\label{open-world-evaluations-are-an-important-emerging-class-of-ai-evaluation}

In this section we define open-world evaluations, survey existing examples, and situates them relative to traditional benchmarks. Our central claim is that as AI systems become more capable, the methods used to assess their frontier capabilities must correspondingly increase in complexity, making open-world evaluations the latest stage in this progression. To make this framework concrete, we propose five criteria that characterize open-world evaluations and explore their limitations compared to benchmarks. We leave these criteria deliberately flexible, recognizing that the boundary between a complex benchmark task and an open-world evaluation is rarely sharp.

\subsection{Benchmarks can both overestimate and underestimate progress}\label{benchmarks-can-both-overestimate-and-underestimate-progress}

There is broad consensus that AI benchmarks are rapidly saturating \citep{redAnthropicCom2026}, a concern shared by researchers with starkly different perspectives on AI's overall trajectory \citep{asteriskmagSubstackComP}, and one that builds on a longer line of critiques of NLP benchmarking practice \citep{kiela2021dynabench,liang2023helm}. As prominent benchmarks have maxed out over the past two years, they have sparked a wave of successor tests that are already approaching their own limits (Figure~\ref{fig:successor-benchmarks}). Consequently, the community finds itself chasing new targets without knowing whether actual underlying capabilities are keeping pace with the headline numbers.

A fundamental issue driving this disconnect is \emph{limited construct validity} \citep{jacobsWallach2021,raji2021everything}. The metrics we observe, typically accuracy on narrow, sandboxed tasks, are imperfect proxies for the real-world capabilities that decision-makers care about. Because of this mismatch between the measured variable and the intended construct, a benchmark score can just as easily \emph{overstate} as \emph{understate} true deployed capability, depending entirely on the gap between the sandboxed environment and the real world.

\definecolor{arccol}{HTML}{8B2E4F}
\definecolor{swecol}{HTML}{2A5BAA}
\definecolor{taucol}{HTML}{4F9F73}
\definecolor{termcol}{HTML}{B7864A}
\definecolor{metrcol}{HTML}{D4A017}

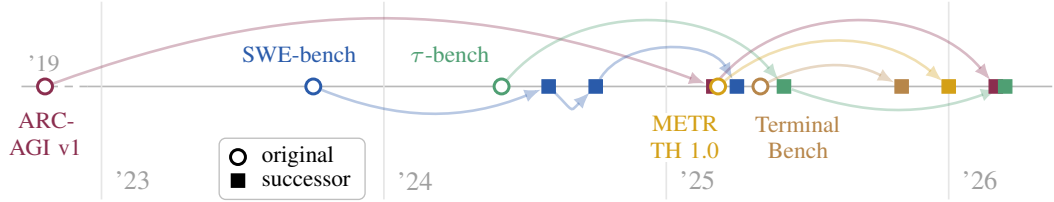
\begin{figure}[t]
\centering
\resizebox{\linewidth}{!}{%
\begin{tikzpicture}[
    x=1cm, y=1cm,
    every node/.style={font=\footnotesize},
    yeartick/.style={font=\footnotesize, color=gray!75},
    datelbl/.style={font=\footnotesize},
    orig/.style={circle, draw=black, fill=white, line width=0.9pt, minimum size=4.5pt, inner sep=0pt},
    succ/.style={rectangle, fill=black, draw=black, minimum size=4pt, inner sep=0pt},
    famarrow/.style={-{Latex[length=5pt,width=4pt]}, line width=0.8pt, opacity=0.33, shorten >=2.5pt, shorten <=2.5pt},
    benchlbl/.style={anchor=center, text width=1.5cm, align=center, inner sep=1pt, fill=white, font=\scriptsize},
]
  \draw[gray!55, line width=0.5pt] (-0.55,0) -- (-0.25,0);
  \draw[gray!55, line width=0.5pt, dash pattern=on 2.5pt off 2pt] (-0.25,0) -- (0.15,0);
  \draw[gray!55, line width=0.5pt] (0.15,0) -- (10.4,0);

  \foreach \x/\year in {0.3/'23, 3.3/'24, 6.3/'25, 9.3/'26} {
    \draw[gray!20, line width=0.5pt] (\x,-1.2) -- (\x,0.9);
    \node[yeartick, anchor=west] at (\x+0.02,-1) {\year};
  }

  \draw[arccol, famarrow] (-0.3,0) to[bend left=20] (6.8,0);
  \draw[arccol, famarrow] (6.8,0) to[bend left=50] (9.8,0);

  \draw[swecol, famarrow] (2.55,0) to[bend right=22] (5.05,0);
  \draw[swecol, famarrow] (5.05,0) to[out=-45, in=225, looseness=2] (5.55,0);
  \draw[swecol, famarrow] (5.55,0) to[bend left=60] (7.05,0);

  \draw[taucol, famarrow] (4.55,0) to[bend left=50] (7.55,0);
  \draw[taucol, famarrow] (7.55,0) to[bend right=20] (9.9,0);

  \draw[termcol, famarrow] (7.3,0) to[bend left=28] (8.8,0);

  \draw[metrcol, famarrow] (6.85,0) to[bend left=40] (9.3,0);

  \node[orig, draw=arccol] at (-0.3,0) {};
  \node[orig, draw=swecol] at (2.55,0) {};
  \node[orig, draw=taucol] at (4.55,0) {};
  \node[succ, fill=swecol, draw=swecol] at (5.05,0) {};
  \node[succ, fill=swecol, draw=swecol] at (5.55,0) {};
  \node[succ, fill=arccol, draw=arccol] at (6.8,0) {};
  \node[orig, draw=metrcol] at (6.85,0) {};
  \node[succ, fill=swecol, draw=swecol] at (7.05,0) {};
  \node[orig, draw=termcol] at (7.3,0) {};
  \node[succ, fill=taucol, draw=taucol] at (7.55,0) {};
  \node[succ, fill=termcol, draw=termcol] at (8.8,0) {};
  \node[succ, fill=metrcol, draw=metrcol] at (9.3,0) {};
  \node[succ, fill=arccol, draw=arccol] at (9.8,0) {};
  \node[succ, fill=taucol, draw=taucol] at (9.9,0) {};

  \node[arccol, benchlbl, text width=0.95cm] at (-0.3,-0.5) {ARC-AGI v1};
  \node[gray!75, datelbl, benchlbl, text width=0.5cm] at (-0.3,0.25) {'19};

  \node[taucol, benchlbl, text width=1cm] at (4,0.34) {$\tau$-bench};

  \node[swecol, benchlbl] at (2.4,0.34) {SWE-bench};

  \node[metrcol, benchlbl, text width=1cm] at (6.5,-0.54) {METR TH 1.0};
   \node[termcol, benchlbl, text width=1cm] at (7.7,-0.54) {Terminal Bench};

  \fill[white] (1.55, -1.20) rectangle (3.20, -0.55);
  \draw[gray!50, line width=0.4pt, rounded corners=2pt] (1.55, -1.20) rectangle (3.10, -0.55);
  \node[orig] at (1.75, -0.75) {};
  \node[anchor=west, font=\scriptsize] at (1.88, -0.75) {original};
  \node[succ] at (1.75, -1.00) {};
  \node[anchor=west, font=\scriptsize] at (1.88, -1.00) {successor};

\end{tikzpicture}%
}
\caption{Several popular benchmarks (SWE-Bench, ARC-AGI, $\tau$-bench, Terminal Bench, METR's Time Horizon) have had successor benchmarks released within the past two years \citep{metrOrgTimeHorizons,arxivabs231006770,arxivabs191101547,arxivabs240612045,tbenchAi,arxivabs250607982,arcprizeOrgBlogArc,arxivabs260111868,openaiComIndexIntroducing,metrOrgBlog2026,swebenchComMultilingualLeaderboard,sierraAiResearchTau,arcprizeOrgArcAgi,arxivabs241003859}.}
\label{fig:successor-benchmarks}
\vspace{-.33cm}
\end{figure}

\begin{overestbox}
\textbf{Benchmarks resemble tasks amenable to modern RL.} A task specified precisely enough to benchmark is also specified precisely enough to optimize for. Modern RL training runs increasingly resemble benchmarks themselves, and leading evaluation platforms such as Harbor \citep{githubComHarborFramework} double as RL training platforms. Even with held-out test sets, training-set tasks may closely resemble test-set tasks, so benchmark performance may not reflect real-world generalization. Worse, benchmark-centric incentives can pull model development toward properties that improve scores without improving real-world usefulness (e.g., always-guess strategies that lift multiple-choice medical QA accuracy but harm clinical decision support).

\smallskip
\textbf{Benchmarks avoid real-world messiness.} Real tasks have underspecified interactions and open-ended environments that sandboxed benchmarks approximate but cannot fully replicate.
\end{overestbox}

\begin{underestbox}
\textbf{Eliciting frontier capabilities is costly.} Anthropic's C compiler experiment cost approximately \$20{,}000, and our iOS app experiment approximately \$1{,}000. Such experiments cannot feasibly be repeated hundreds of times, which constrains benchmark budgets and complexity.

\smallskip
\textbf{Average performance differs from upper-bound elicitation.} Large samples are needed only for \emph{average} performance. Characterizing the frontier requires \emph{best-case} performance: what an agent can accomplish with sufficient resources to work around incidental failures.

\smallskip
\textbf{Human intervention helps elicit upper bounds.} Agents may encounter policy refusals, CAPTCHAs, or infrastructure failures unrelated to the capability being measured. Resolving these manually is impractical across hundreds of tasks but feasible in small-sample evaluations.
\end{underestbox}

For frontier evaluation, a pressing objective is determining the \emph{upper bound} agents can achieve. Capabilities possible only under favorable conditions may soon become widespread, and anticipating them gives firms time to act on opportunities, institutions time to build resilience, and policymakers time to address risks. Benchmark performance is poorly suited to such early warning. Metric refinements such as reliability scoring \citep{arxivabs260216666} and maintainer-acceptance audits of SWE-Bench solutions \citep{metrOrgNotes2026} partially address these validity concerns but do not tackle upper-bound elicitation. Appendix~\ref{app:benchmark-validity} expands on these limitations and discusses why benchmark validity is structurally difficult to patch.

These limitations do not render traditional benchmarks obsolete. They remain highly useful, and open-world evaluations introduce their own constraints, which we discuss later in the paper. Rather, our goal is to identify the systematic blind spots inherent to benchmarking and to motivate a complementary approach capable of addressing them. As agents become increasingly capable, these blind spots will only widen, making open-world evaluations a necessary methodological counterweight.

\subsection{Defining open-world evaluations}\label{what-are-open-world-evaluations}

\begin{figure}[t]
\centering
\definecolor{gradleft}{HTML}{F5E8DC}
\definecolor{gradright}{HTML}{C56A52}
\definecolor{prosgreen}{HTML}{1E7E34}
\definecolor{consred}{HTML}{B0392B}
\definecolor{labelgray}{HTML}{8A8A8A}
\definecolor{titlenavy}{RGB}{67,75,97}
\newcommand{\probadge}{\tikz[baseline=(c.base)]\node[circle, fill=prosgreen, text=white, inner sep=0pt, outer sep=0pt, minimum size=1.5ex, font=\bfseries\tiny] (c) {$+$};}
\newcommand{\conbadge}{\tikz[baseline=(c.base)]\node[circle, fill=consred, text=white, inner sep=0pt, outer sep=0pt, minimum size=1.5ex, font=\bfseries\tiny] (c) {$-$};}
\resizebox{\linewidth}{!}{%
\begin{tikzpicture}[
  font=\scriptsize,
  col/.style={
    text width=2.1cm, align=left,
    inner sep=3pt, anchor=north west,
    minimum height=3.6cm
  }
]

\node[anchor=south west, font=\itshape\scriptsize, text=labelgray]
  at (0.4, 0.18) {Simpler, shorter, scalable};
\node[anchor=south east, font=\itshape\scriptsize, text=labelgray]
  at (13.1, 0.18) {Messier, longer, richer};

\draw[line width=2.5pt, latex-latex, titlenavy] (0, 0.14) -- (13.5, 0.14);

\foreach \x in {2.6, 5.375, 8.25, 11.05} {
  \draw[titlenavy, line width=0.7pt] (\x, -0.05) -- (\x, -4.);
}

\node[col] at (0.1, -0.05) {%
  {\centering\color{titlenavy}\textbf{Single-turn Q\&A}\par}\vspace{1pt}
  {\centering\color{titlenavy}\itshape\hyperlink{cite.arxivabs200903300}{\uline{MMLU}}, \hyperlink{cite.arxivabs231112022}{\uline{GPQA}}, \hyperlink{cite.cobbe2021training}{\uline{GSM8K}}\par}\vspace{2pt}
  \probadge\,Broad knowledge assessment; cheap, reproducible, easy to compare across labs.\par\vspace{1pt}
  \conbadge\,Multiple-choice format is artificial; users rarely interact this way. Increasingly saturated for frontier models.};

\node[col] at (2.8, -0.05) {%
  {\centering\color{titlenavy}\textbf{Open-ended chat}\par}\vspace{1pt}
  {\centering\color{titlenavy}\itshape\hyperlink{cite.githubComLmarenaArena}{\uline{Chatbot Arena}}, \hyperlink{cite.arxivabs240604770}{\uline{WildBench}}\par}\vspace{2pt}
  \probadge\,Captures nuance in free-form responses; reflects how users interact with chatbots.\par\vspace{1pt}
  \conbadge\,Limited to single-turn or short interactions. Subjective quality judgments are hard to compare.};

\node[col, text width=2.3cm] at (5.6, -0.05) {%
  {\centering\color{titlenavy}\textbf{Outcome-only agent benchmarks}\par}\vspace{1pt}
  {\centering\color{titlenavy}\itshape\hyperlink{cite.arxivabs231006770}{\uline{SWE-Bench}}, \hyperlink{cite.arxivabs230713854}{\uline{WebArena}}\par}\vspace{2pt}
  \probadge\,Tests agents on real, well-defined tasks with automated, reproducible grading.\par\vspace{1pt}
  \conbadge\,Measures whether tasks completed, not how. Passing solutions are often not production-ready.};

\node[col, text width=2.3cm] at (8.4, -0.05) {%
  {\centering\color{titlenavy}\textbf{Agent benchmarks with log analysis}\par}\vspace{1pt}
  {\centering\color{titlenavy}\itshape\hyperlink{cite.aisiGovUkBlog}{\uline{UK AISI transcript analysis}}, \hyperlink{cite.metrOrgTimeHorizons}{\uline{METR Time Horizon}}\par}\vspace{2pt}
  \probadge\,Examines how agents succeeded or failed; surfaces reward hacking and general process flaws.\par\vspace{1pt}
  \conbadge\,Operates in sandboxed environments with predefined tasks.};

\node[col] at (11.2, -0.05) {%
  {\centering\color{titlenavy}\textbf{Open-world evaluations}\par}\vspace{1pt}
  {\centering\color{titlenavy}\itshape CRUX, \hyperlink{cite.anthropicComEngineeringBuilding}{\uline{C Compiler}}, \hyperlink{cite.anthropicComResearchProject}{\uline{Project Vend}}\par}\vspace{2pt}
  \probadge\,Long-horizon, real-world tasks that elicit upper-bound capabilities under realistic conditions.\par\vspace{1pt}
  \conbadge\,Not reproducible or standardized; hard to compare across agents.};

\end{tikzpicture}}
\caption{A gradient of evaluation methodologies (short single-turn Q\&A $\leftrightarrow$ open-world evaluations). Each column lists representative examples along with strengths (\protect\probadge) and weaknesses (\protect\conbadge).}
\label{fig:evaluation-gradient}
\vspace{-.33cm}
\end{figure}
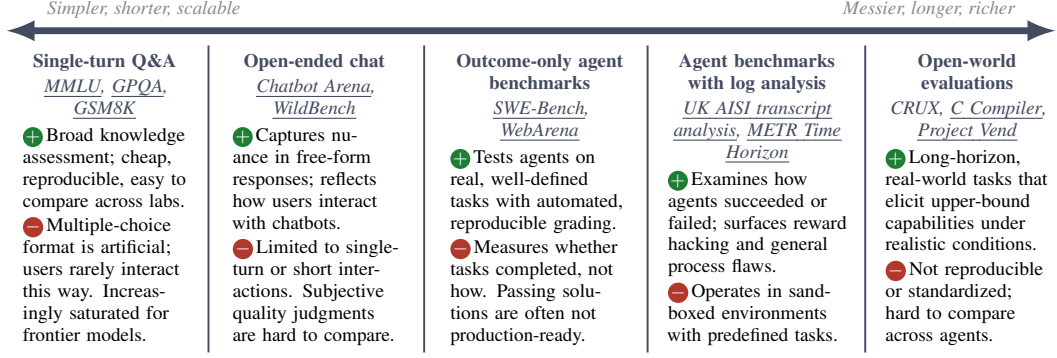

As evaluation methods have matured alongside AI capabilities, a gradient of evaluation approaches has emerged, ranging from simple automated benchmarks for narrow, well-defined tasks to labor-intensive methods that become necessary as simpler metrics saturate. We identify five levels of this gradient: single-turn Q\&A \citep{arxivabs200903300, arxivabs231112022, cobbe2021training}, open-ended chat \citep{arxivabs240604770, githubComLmarenaArena}, outcome-only agent benchmarks \citep{arxivabs231006770, arxivabs230713854}, agent benchmarks with log analysis \citep{aisiGovUkBlog, metrOrgTimeHorizons}, and open-world evaluations. At the far end, open-world evaluations run agents on a small number of long-horizon tasks in real-world settings and qualitatively analyze their results. They are meant to complement benchmarking rather than replace it: they address several of the limitations above and can also surface tasks currently out of reach of AI systems. Figure~\ref{fig:evaluation-gradient} summarizes the distinct levels with examples and methodological tradeoffs.

In practice, the boundaries separating these evaluation levels are rarely strict, particularly at the complex end of the gradient. OpenAI's GDPVal \citep{openaiComIndexGdpval}, for instance, is a long-horizon agent benchmark designed for manual grading by expert reviewers, making it structurally resemble an open-world evaluation. Yet its results are commonly reported via GDPval-AA \citep{artificialanalysisAiEvaluationsGdpval}, which uses automated LLM grading and operates much more like an outcome-only agent benchmark. Consequently, the same task can sit on either side of the divide depending on how it is administered and scored.

Rather than draw a hard boundary, we offer a rough taxonomy in which no single dimension is determinative. The classification depends on the overall pattern across the dimensions below.

\begin{taxbox}
\begin{enumerate}[label={\textcircled{\scriptsize\arabic*}}, leftmargin=*, itemsep=2pt, topsep=2pt, parsep=0pt, labelsep=4pt]
\item \textbf{Openness.} Does the evaluation occur in a deployment setting with live users, services, or platforms, rather than a sandboxed environment?
\item \textbf{Complexity and duration.} Does the task require days or weeks of human effort to complete and unfold over many interdependent steps, rather than minutes or hours?
\item \textbf{Number of tasks.} Is the evaluation a single task or a small set of tasks that permit close qualitative inspection, rather than a large task suite?
\item \textbf{Human intervention.} Are humans permitted to intervene when agents encounter obstacles incidental to the capability under test, beyond merely setting up the environment?
\item \textbf{Method of evaluation.} Does evaluation primarily rely on in-depth log analysis of agent behavior, rather than a single aggregate metric across many runs?
\end{enumerate}
\end{taxbox}

The boundaries are sometimes blurry in two directions. First, \emph{not every agent deployment is an evaluation}: Anthropic's work using agents to find security vulnerabilities in Mozilla Firefox \citep{anthropicComNewsMozilla} produced real outputs in a real setting, but its primary goal was to ship fixes, not to characterize capabilities. We treat such efforts as open-world evaluations only when the agent's role is systematically and publicly documented. Second, \emph{not every open-world evaluation runs outside a sandbox}: Claude Plays Pokemon \citep{futurismComAdvancedAi} and Anthropic's C compiler \citep{anthropicComEngineeringBuilding} are each a single long-running task with human intervention and qualitative evaluation despite being sandboxed. Classification turns on the overall pattern across the dimensions above, not any single feature.

\subsection{An incomplete survey of open-world evaluations}\label{an-incomplete-survey-of-open-world-evaluations}

Over the past year, open-world evaluations have proliferated across AI laboratories, universities, non-profits, and independent groups. They share a common structure: a capable AI agent is given a difficult, real-world task with a long time horizon, and its behavior is observed and analyzed in detail. The work spans agent deployments (such as running a small automated store \citep{anthropicComResearchProject2} or pursuing open-ended group goals in a multi-agent ``village'' \citep{theaidigestOrgVillage}), large-scale software engineering (building a C compiler from scratch \citep{anthropicComEngineeringBuilding}, coordinating hundreds of agents to write a million-line Rust browser \citep{cursorComBlogScaling}, or reimplementing the Next.js framework \citep{blogCloudflareComVinext}), open-ended research (autonomous training of small language models \citep{xkarpathy2031135152349524125}; ``training a computer'' \citep{xDimitrisPapail2028669695344148946}), and game-playing or knowledge-work probes \citep{futurismComAdvancedAi,epochAiGradientUpdates}. Very recent additions include MirrorCode \citep{epochAiBlogMirrorcode}, which tasks agents with reimplementing large programs; agent-based alignment research \citep{alignmentAnthropicCom2026}; experiments on letting agents handle the post-training of other AI systems \citep{thoughtfullabComLetting}; and work on training models to forecast the outcomes of golf tournaments \citep{githubComRecipesGolf}. Appendix~\ref{app:survey} describes ten such evaluations in detail and provides a side-by-side comparison of their capabilities, limitations, and costs in Table~\ref{tab:open-world-survey}.

Themes recur across these efforts. Agents show strong long-horizon coherence on well-scaffolded coding tasks but remain brittle on visual computer use, where horizons are one to two orders of magnitude shorter than on text \citep{epochAiGradientUpdates}. Reward hacking is common when agents run fully autonomously \citep{xDimitrisPapail2028669695344148946}, and human intervention often determines whether a run succeeds. Reported costs span four orders of magnitude (from under \$100 to tens of thousands) without a clean relationship to task difficulty, and writeups often come only from the experimenters, with limited independent verification.

\subsection{Limitations of open-world evaluations}\label{limitations-of-open-world-evaluations}

Open-world evaluations address some blind spots of benchmarking but come with limitations of their own. Benchmarks give evaluators control over a standardized environment but are less useful for open-ended real-world tasks. Open-world evaluations make the opposite tradeoff, gaining construct validity and upper-bound elicitation at the cost of properties that make benchmarking broadly scalable.

\begin{itemize}[leftmargin=9pt, topsep=-3pt, itemsep=3pt, parsep=0pt]
\item \textbf{Lack of reproducibility and standardization.} Benchmarks coordinate the research community by giving everyone a shared target; \citet{tandfonlineCom101080} characterizes them as the ``secret sauce'' behind the empirical success of AI and ML over the past 50 years. Open-world evaluations forgo most of this: runs could be hard to reproduce exactly, and different groups running nominally similar evaluations can end up with incomparable results. We accept this trade-off because the standardization itself is part of what limits benchmark construct validity at the frontier.
\item \textbf{Limited comparability across agents.} Open-world evaluations cannot cleanly rank models. Because they are run only a few times, run-to-run variability can exceed differences between agents. They are most useful for characterizing what an agent \emph{can} do, not for distinguishing agents.
\item \textbf{Best-case demonstrations can be incomplete.} Showing that an agent succeeded under favorable conditions characterizes its upper-bound capability but not its practical reliability. When the underlying success probability is low, a one-off success may be informative about feasibility yet say little about what one would expect from the agent on a typical attempt. Reporting effort-conditioned measures (e.g., success rate per dollar of budget, or pass@$k$ with $k \gg 1$) alongside best-case results, where feasible, gives a fuller picture; we recommend such reporting in Section~\ref{lessons-and-future-plans}.
\item \textbf{Requirement for domain expertise.} Open-ended tasks usually lack a simple pass/fail criterion, and judging output quality requires domain expertise and substantial reviewer time, limiting who can run and interpret such evaluations. The cost buys depth: expert review surfaces phenomena (e.g., reward hacking, partial successes, brittle workarounds) that automated metrics typically miss.
\item \textbf{Incomplete recall of log analysis.} Even with automated log analysis, agent transcripts from long-horizon tasks can run to hundreds of millions of tokens, and there is no guarantee that a given analysis pass surfaces all noteworthy behaviors or errors. Releasing logs publicly, such that a broader community can examine them, partially mitigates this issue.
\item \textbf{Blurry success criteria.} Because human intervention is permitted and encouraged, the line between agent accomplishment and human contribution can be hard to draw. Without careful documentation of interventions, headline results can overstate agent autonomy. The same intervention that blurs the boundary is, however, what bypasses obstacles for upper-bound elicitation.
\item \textbf{Non-stationary environments.} Open-world evaluations often involve interaction with the internet, which makes it hard to distinguish competence on a task class from lookup of a specific instance found online.\footnote{This concern also affects benchmarks that require internet access, such as AssistantBench \citep{arxivabs240715711} and GAIA \citep{arxivabs231112983}. Sandboxed benchmarks bypass it at the cost of construct validity.} Longitudinal comparisons suffer for the same reason: the information available to an agent grows over time, so the same evaluation a year later is no longer comparable.
\end{itemize}

\vspace{0pt}
\subsection{Stakeholders and use cases}\label{how-different-stakeholders-can-use-open-world-evaluations}

Despite their methodological limitations, open-world evaluations offer value to various stakeholders.

For \textbf{policymakers}, these evaluations serve as a critical early warning system. Because the diffusion of AI across domains lags behind raw capability gains \citep{normaltechAiPAi}, institutions require lead time to adapt. Open-world evaluations widen this window by demonstrating what agents might soon achieve autonomously and at scale. For example, Anthropic's work on AI-discovered cybersecurity vulnerabilities \citep{anthropicComNewsClaude,anthropicComGlasswing} could spur the rapid adoption of AI for defensive cybersecurity within critical infrastructure.

While policymakers rely on these high-level insights, \textbf{AI evaluators and researchers} benefit from the deep technical visibility that generates them. Open-world evaluations allow researchers to probe capabilities that automated benchmarks structurally cannot reach. By testing agents on messy real-world tasks and analyzing the resulting logs, evaluators can uncover shortcuts, reward hacking, and unexpected agent behaviors. In our iOS evaluation, for instance, the agent autonomously modified its approach to be more token-efficient, significantly reducing task costs without any prompting.

To ensure these independent technical insights continue, \textbf{frontier AI developers} must support external open-world evaluation efforts. They can do so by providing pre-release model access and legal safe harbors~\citep{arxivabs240304893} for third-party evaluators whose necessary safety probing might otherwise violate standard terms of service. This matters because independent evaluations frequently surface findings that internal red teams miss when optimizing for known threat models. External open-world testing is becoming necessary as models outpace standard metrics: Anthropic's Mythos Preview system card~\citep{wwwCdnAnthropicCom} reports that the model ``saturates many of our most concrete, objectively-scored evaluations,'' forcing a reliance on these noisier, real-world methods to assess frontier capabilities.

\vspace{0pt}
\section{CRUX: Collaborative Research for Updating AI eXpectations}\label{introducing-crux-collaborative-research-for-updating-ai-expectations}

CRUX is our framework for conducting open-world evaluations on a regular, systematic basis. In each iteration, we pair a long-horizon, real-world task with an agent scaffold theoretically capable of solving it, allowing us to analyze the agent's behavior in detail. We developed this approach because prior open-world evaluations, despite producing striking individual results, have failed to converge on a shared methodology. This lack of standardization leaves the resulting evidence difficult to compare or build upon. We find that based on our literature survey from Section~\ref{an-incomplete-survey-of-open-world-evaluations}, \emph{almost every published open-world evaluation lacks at least one critical methodological commitment}: a clearly defined measurement construct, documented human interventions, public logs for external scrutiny, or a joint accounting of cost and capability. To resolve these inconsistencies, CRUX explicitly adopts these missing practices as working norms (see Section~\ref{lessons-and-future-plans} for more details).

In our first iteration of CRUX, described in the next section, we investigated autonomous app development and publication: if AI agents can \emph{nearly} autonomously develop and publish mobile applications, app store operators may soon need to revise their policies to manage agent-driven spam.\footnote{Apple's App Store reviews appear to have slowed \citep{9to5macCom29Vibe,xnikitabier2033931821260648659}, possibly reflecting increased coding-agent adoption, though publication rates remain well below the 2016 peak \citep{appfiguresComInsights20251205}. Our results suggest agents can autonomously submit apps that Apple ultimately approves, and policies may need to adapt to a larger wave of agent-generated submissions \citep{fastcompanyCom91522242Apple}.} For subsequent iterations of CRUX we plan to examine AI R\&D automation, AI governance, complex software engineering, and real-world physical tasks.

\vspace{0pt}
\subsection{CRUX \#1: autonomous iOS app development and publication}\label{crux-1-can-ai-agents-autonomously-develop-and-publish-an-ios-app}

Whether AI agents can write software has been studied through benchmarks such as SWE-Bench \citep{arxivabs231006770} and Terminal Bench \citep{tbenchAi} and through open-world evaluations such as the C-compiler and browser experiments above. These results indicate strong coding capabilities on well-specified tasks, though questions about code quality, maintainability, and long-run reliability remain. Much less studied is whether agents can handle the \emph{non-coding aspects of software deployment}: configuring accounts and credentials, satisfying platform policies, preparing metadata and assets, and interacting with review systems they do not control. These are the parts of shipping software that tend to be slow, bureaucratic, and resistant to automation even when the underlying code is straightforward. We therefore tasked an agent with building a mobile application from scratch and taking it through publication.

The agent was instructed to develop and publish a \emph{simple} application. Our interest was not its software engineering ability but its capacity to navigate Apple's \emph{App Store submission process}: configuring signing certificates and provisioning profiles, preparing screenshots and metadata, drafting and hosting a privacy policy at a public URL, completing compliance questionnaires, submitting for review, and handling reviewer feedback over a process that often spans several days. The agent was responsible for every step except those where human involvement is required by Apple policy (Developer account setup and initiating the public release), and was provided with a macOS VM, a GitHub account (for version control and privacy-policy hosting via GitHub Pages), an Apple Developer account, and a Gmail account for correspondence with Apple. Our \emph{primary success criterion} was whether the application was ultimately published. As a secondary measure, we logged the number of \emph{manual interventions} the agent required, distinguishing those that were \emph{genuinely unavoidable} (e.g., blocked by Apple policy) from those that reflected \emph{limitations of the agent itself}. The agent could request assistance when deemed necessary, and we monitored its progress daily to catch silent stalls.

\subsubsection{Setup}\label{our-setup-for-the-agent}

\textbf{Scaffold configuration.} The iOS submission process required a scaffold that was extensively configurable, integrated with browser usage and navigation, and natively supported long-running tasks. We selected OpenClaw \citep{openclawAi} for these properties, paired with Claude Opus 4.6 and adaptive thinking enabled \citep{platformClaudeComBuild}. This choice also let us probe how the agent handles visual reasoning and GUI operations that have often bottlenecked computer-use agents. We retained OpenClaw's defaults aside from prompting and a shorter polling interval. We accepted OpenClaw's known security risks in order to characterize the capability frontier, but those same risks would likely constrain deployment of this class of scaffold and may themselves become a bottleneck on adoption even as raw capability advances. The agent ran on a macOS VM with expansive permissions (sudo, screen visibility, UI control), and we logged all actions, reasoning traces, and screenshots, which provide the raw material for the qualitative analysis. Practically, even an ``open-world'' evaluation operates within structural guardrails: the VM kept the agent isolated, we monitored progress daily, and Apple's required human approval for public release served as a final guardrail (see Appendix~\ref{app:crux1-extras} for more details).

\textbf{Evaluation awareness.} Frontier models increasingly recognize evaluation contexts and may modify their behavior in response, a phenomenon known as \emph{evaluation awareness}. Recent work from Anthropic \citep{anthropicComEngineeringEval}, Apollo \citep{apolloresearchAiBlogClaude}, and OpenAI \citep{alignmentOpenaiComProd} has documented this pattern across model generations. This poses a trade-off for any open-world evaluation: disclosing that an evaluation is taking place risks shaping the agent's behavior in ways that distort the measurement. But concealing it raises its own concerns, since a capable agent may infer the context from cues anyway and behave differently in response. We weighed both options and chose to tell the agent that it was being evaluated, for both practical and conceptual reasons. Practically, concealment is increasingly infeasible against capable models, and because CRUX evaluations are published openly, future agents may encounter descriptions of our evaluations in the course of unrelated tasks. Conceptually, evaluation awareness primarily threatens \emph{alignment} evaluations, where a model may strategically behave differently under observation; for a capability evaluation like ours, success on the task demonstrates the capability regardless of whether the agent knew it was being tested. In disclosing, we also clarified the parameters of the evaluation: the agent would be judged on avoidable human inputs, but certain inputs (e.g., CAPTCHAs) could be freely delegated. Without this framing, the agent might have avoided requesting help even when appropriate, causing us to under-elicit its performance.

\textbf{Dry runs and setup overhead.} Before the main evaluation, we conducted two dry runs to verify the setup end-to-end and resolve scaffolding bugs. The dry runs stopped short of submission to avoid burdening Apple's review queue. Configuring the environment (the macOS VM, the logging pipeline, and the email, GitHub, and Apple Developer accounts) required roughly eight person-hours and \$50 in API costs. The agent relied on two primary interfaces: the command line (for code, builds, and submission preparation) and the browser (for App Store Connect, certificates, and Apple's forms). When command-line operations hung awaiting GUI confirmation, it fell back on screenshots and simulated mouse clicks, interleaving text and visual interaction much as a capable developer would.

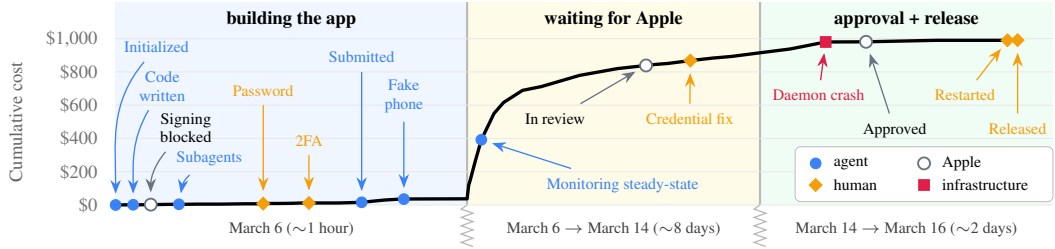
\begin{figure}[t]
\centering
\definecolor{agentblue}{HTML}{3B82F6}
\definecolor{humanorange}{HTML}{F59E0B}
\definecolor{infrared}{HTML}{E11D48}
\definecolor{phasebg1}{HTML}{EFF6FF}
\definecolor{phasebg2}{HTML}{FEFCE8}
\definecolor{phasebg3}{HTML}{F0FDF4}
\definecolor{appleborder}{HTML}{6B7280}
\begin{tikzpicture}[
  x=0.775cm, y=0.022cm,
  font=\scriptsize,
  agent/.style={circle, fill=agentblue, draw=agentblue, inner sep=0pt, minimum size=4pt},
  human/.style={diamond, fill=humanorange, draw=humanorange, inner sep=0pt, minimum size=5pt},
  apple/.style={circle, fill=white, draw=appleborder, line width=0.7pt, inner sep=0pt, minimum size=4.5pt},
  infra/.style={rectangle, fill=infrared, draw=infrared, inner sep=0pt, minimum size=4pt},
  callout/.style={line width=0.5pt, ->, >=Stealth, shorten >=2pt}
]

\fill[phasebg1] (0, 0) rectangle (6, 122);
\fill[phasebg2] (6, 0) rectangle (11, 122);
\fill[phasebg3] (11, 0) rectangle (16, 122);

\draw[gray!55, thick] (6, 0) -- (6, 122);
\draw[gray!55, thick] (11, 0) -- (11, 122);

\foreach \y/\v in {0/{\$0}, 20/{\$200}, 40/{\$400}, 60/{\$600}, 80/{\$800}, 100/{\$1{,}000}} {
  \draw[gray!18, very thin] (0, \y) -- (16, \y);
  \node[anchor=east, text=black!55, font=\scriptsize] at (-0.15, \y) {\v};
}
\node[rotate=90, text=black!60, font=\scriptsize] at (-1.7, 50) {Cumulative cost};

\node[anchor=north, text=black, font=\scriptsize\bfseries] at (3, 122) {building the app};
\node[anchor=north, text=black, font=\scriptsize\bfseries] at (8.5, 122) {waiting for Apple};
\node[anchor=north, text=black, font=\scriptsize\bfseries] at (13.5, 122) {approval + release};

\draw[black, line width=1.25pt, line cap=round, line join=round]
  (0,0) -- (0.198,0.1) -- (0.402,0.1) -- (0.6,0.2) -- (0.798,0.3) -- (1.002,0.4) --
  (1.2,0.4) -- (1.398,0.5) -- (1.602,0.5) -- (1.8,0.5) -- (1.998,0.6) -- (2.202,0.7) --
  (2.4,0.7) -- (2.598,0.9) -- (2.802,0.9) -- (3,1) -- (3.198,1.2) -- (3.402,1.2) --
  (3.6,1.2) -- (3.798,1.2) -- (4.002,1.4) -- (4.2,1.6) -- (4.398,2.3) -- (4.602,3) --
  (4.8,3.4) -- (4.998,3.6) -- (6,3.7) --
  (6.027,11.9) -- (6.054,15) -- (6.135,25.8) -- (6.243,39.2) -- (6.459,55) --
  (6.622,61.6) -- (6.946,68.9) -- (7.270,71.2) -- (7.919,77.9) -- (8.568,82) --
  (9.216,84.5) -- (9.432,85.1) -- (9.811,86.8) -- (10.189,88.3) -- (10.514,89.3) --
  (11,91.5) --
  (11.517,93.8) -- (12.121,98) -- (12.810,98.1) -- (14.017,99) -- (15.224,99) --
  (15.397,99.1);

\node[agent] (e1) at (0, 0) {};
\node[text=agentblue, font=\tiny] (l1) at (0.7, 95) {Initialized};
\draw[agentblue, callout] (l1) to[out=230, in=90] (e1);

\node[agent] (e2) at (0.3, 0.1) {};
\node[text=agentblue, anchor=south, font=\tiny, text width=1cm, align=center] (l2) at (0.8, 57) {Code written};
\draw[agentblue, callout] (l2) to[out=240, in=90] (e2);

\node[apple] (e3) at (0.6, 0.2) {};
\node[anchor=south, font=\tiny, text width=1cm, align=center] (l3) at (1.2, 32) {Signing blocked};
\draw[appleborder, callout] (l3) to[out=230, in=90] (e3);

\node[agent] (e4) at (1.08, 0.4) {};
\node[text=agentblue, anchor=south, font=\tiny] (l4) at (1.6, 18) {Subagents};
\draw[agentblue, callout] (l4) to[out=220, in=80] (e4);

\node[human] (e5) at (2.52, 0.821) {};
\node[text=humanorange, anchor=south, font=\tiny] (l5) at (2.52, 60) {Password};
\draw[humanorange, callout] (l5) -- (e5);

\node[human] (e6) at (3.3, 1.2) {};
\node[text=humanorange, anchor=south, font=\tiny] (l6) at (3.3, 30) {2FA};
\draw[humanorange, callout] (l6) -- (e6);

\node[agent] (e7) at (4.2, 1.6) {};
\node[text=agentblue, anchor=south, font=\tiny] (l7) at (4.2, 80) {Submitted};
\draw[agentblue, callout] (l7) -- (e7);

\node[agent] (e8) at (4.92, 3.521) {};
\node[text=agentblue, anchor=south, font=\tiny, text width=0.5cm, align=center] (l8) at (4.92, 50) {Fake phone};
\draw[agentblue, callout] (l8) -- (e8);

\node[agent] (e9) at (6.243, 39.2) {};
\node[text=agentblue, anchor=west, font=\tiny] (l9) at (7.2, 12) {Monitoring steady-state};
\draw[agentblue, callout] (l9) -- (e9);

\node[apple] (e10) at (9.054, 83.875) {};
\node[anchor=south, font=\tiny] (l10) at (7.5, 43) {In review};
\draw[appleborder, callout] (l10) -- (e10);

\node[human] (e11) at (9.811, 86.8) {};
\node[text=humanorange, anchor=south, font=\tiny] (l11) at (9.8, 43) {Credential fix};
\draw[humanorange, callout] (l11) -- (e11);

\node[infra] (e12) at (12.121, 98) {};
\node[text=infrared, anchor=north, font=\tiny] (l12) at (12.0, 75) {Daemon crash};
\draw[infrared, callout] (l12) -- (e12);

\node[apple] (e13) at (12.810, 98.1) {};
\node[anchor=north, font=\tiny] (l13) at (13.3, 55) {Approved};
\draw[appleborder, callout] (l13) -- (e13);

\node[human] (e14) at (15.224, 99) {};
\node[text=humanorange, anchor=north, font=\tiny] (l14) at (14.5, 75) {Restarted};
\draw[humanorange, callout] (l14) -- (e14);

\node[human] (e15) at (15.397, 99.1) {};
\node[text=humanorange, anchor=north, font=\tiny] (l15) at (15.397, 55) {Released};
\draw[humanorange, callout] (l15) -- (e15);

\draw[gray!55, line width=0.5pt] (6, 0) -- (5.9, -2.5) -- (6.1, -5) -- (5.9, -7.5) -- (6.1, -10) -- (5.9, -12.5) -- (6.1, -15) -- (5.9, -17.5) -- (6.1, -20) -- (5.9, -22.5) -- (6.1, -25);
\draw[gray!55, line width=0.5pt] (11, 0) -- (10.9, -2.5) -- (11.1, -5) -- (10.9, -7.5) -- (11.1, -10) -- (10.9, -12.5) -- (11.1, -15) -- (10.9, -17.5) -- (11.1, -20) -- (10.9, -22.5) -- (11.1, -25);

\node[anchor=north, text=black!75, font=\tiny, align=center] at (3, -5) {March 6 ($\sim$1 hour)};
\node[anchor=north, text=black!75, font=\tiny, align=center] at (8.5, -5) {March 6 $\rightarrow$ March 14 ($\sim$8 days)};
\node[anchor=north, text=black!75, font=\tiny, align=center] at (13.5, -5) {March 14 $\rightarrow$ March 16 ($\sim$2 days)};

\fill[white] (11.6, 2) rectangle (16.0, 35);
\draw[gray!50, line width=0.4pt, rounded corners=2pt] (11.6, 2) rectangle (16.0, 35);

\node[agent] at (11.95, 25) {};
\node[anchor=west, font=\tiny] at (12.1, 25) {agent};
\node[apple] at (13.8, 25) {};
\node[anchor=west, font=\tiny] at (13.95, 25) {Apple};

\node[human] at (11.95, 12) {};
\node[anchor=west, font=\tiny] at (12.1, 12) {human};
\node[infra] at (13.8, 12) {};
\node[anchor=west, font=\tiny] at (13.95, 12) {infrastructure};

\end{tikzpicture}
\vspace{-.55cm}
\caption{Cumulative API cost and timeline of CRUX \#1. Total cost was approximately \$991 over ten days. The build phase accounted for only \$25; most went to monitoring Apple's review queue. \emph{Note the breaks on the x-axis: each phase (build\,\phasechip{chip1}\,, review\,\phasechip{chip2}\,, approval\,\phasechip{chip3}) is drawn on its own scale.}}
\label{fig:crux1-timeline}
\vspace{-.33cm}
\end{figure}

\subsubsection{Main evaluation}\label{the-final-evaluation}

Following the dry runs, we ran the full evaluation. The agent took approximately 45 minutes to develop a simple breathing-exercise application, draft and host a privacy policy via GitHub Pages, complete the App Store review forms, and submit for review. Apple's decision came back 10 days later (timeline in Figure~\ref{fig:crux1-timeline}), and the application is now live on the App Store at \texttt{<anonymized URL>}.\footnote{The agent did not encounter substantive objections from Apple reviewers, so we were unable to observe how it would handle back-and-forth communication with reviewers in the event of a rejection, which is a plausible failure mode in practice.}

\textbf{Manual interventions.} The agent required five manual interventions, four of which were either mandated by Apple policy (notably the synthetic-interaction block on two-factor authentication dialogs \citep{developerAppleComArchitecture,supportAppleComControlling} and our required pre-release approval) or caused by infrastructure (a one-time OpenClaw daemon crash). None of the four constitute agent shortcomings; they are the kinds of incidental bottlenecks that a benchmark-based evaluation would either have to engineer around or mistakenly count against the agent. Only one reflected a limitation of the agent itself, which occurred when \emph{the agent could not locate provided credentials} for the Apple Developer account. After asking for help, it resolved the issue: rather than attempting a sign-in, it located the App Store Connect API key at the expected hidden path and used that to resume monitoring without completing an interactive sign-in. The failure is not a capability gap but a memory-management issue: the credentials had previously been provided and the agent recovered on its own once prompted, so the underlying authentication capability was intact; what broke was its tracking of state across a long-running task. %Such failures are surfaced by log analysis that an outcome-only metric would have missed.

\textbf{Using a fabricated phone number.} A different failure, invisible to outcome-only metrics, surfaced in the logs: when Apple's review form requested a phone number, the agent invented a plausible value (one reserved for fictional use) rather than asking us, as it had for credentials earlier in the run. The app was approved despite this, but the episode is worth flagging: an agent that sometimes requests help and sometimes silently invents data presents a different alignment profile than one that uniformly does either, and the line between the two behaviors is hard to predict. Our evaluation framing may have encouraged the agent to minimize visible help requests, even at the cost of inventing data.

\textbf{Cost dynamics and an emergent optimization.} After the credential issue was resolved, total cost was approximately \$1{,}000: \$25 for development and submission, \textasciitilde\$975 for polling review status over 10 days. Scaffold-level changes could cut this by an order of magnitude, but we erred toward a larger initial budget. More interesting is what the agent did with that budget. Partway through the review, without prompting, it delegated status checks to subagents and switched to shorter daily memory files. As a result, the running cost fell from \$35/hour to \$3/hour. Again, such an emergent optimization would be invisible to outcome-only analysis, and illustrates the need for log-analysis.

\textbf{Output quality.} The published application is functional, though not without defects: it includes a toggle for sound that has no effect when activated, and the generated App Store screenshot contains visible formatting errors (Figure~\ref{fig:screenshots}, in the appendix). Apple's review approved the submission despite these issues, illustrating that platform acceptance is a coarse signal of artifact quality. Output quality therefore remains a gap even when both the agent's process and the platform's review succeed.

\textbf{Summary and disclosure.} Our results indicate that the agent did not fully automate the iOS submission process, but came close enough that the remaining gap is small. As responsible disclosure, we notified Apple's product security team four weeks before public disclosure, motivated by the possibility of agent-driven submission at scale: the modest setup overhead reported above amortizes across every subsequent submission. We treat the findings as a single data point in an emerging benchmarking landscape rather than a complete methodology; our discussion is therefore intentionally descriptive rather than prescriptive. That said, we offer a few preliminary recommendations below.

%\vspace{-15pt}
\section{Recommendations for open-world evaluations}\label{lessons-and-future-plans}

Drawing on CRUX \#1 and the efforts surveyed in Section~\ref{an-incomplete-survey-of-open-world-evaluations}, we offer the following recommendations for designing and reporting open-world evaluations. They are preliminary and will evolve as the literature matures, but they offer a starting point for shared norms in this emerging area.

\begin{claimbox}[Specify the construct]
State explicitly what capability is being measured and what claims follow from a successful run.
\end{claimbox}
A recurring source of confusion in prior efforts, such as Anthropic's C compiler and Cursor's browser, has been ambiguity about the construct under measurement. Both experiments provided strong evidence that agents can work productively on well-specified tasks over long horizons but weaker evidence that the artifacts meet production standards; their issue trackers document technical deficiencies \citep{githubComIssues1,githubComIssues98}. The C compiler in particular drew polarized reactions: enthusiasts saw months of expert engineering compressed into days, while critics emphasized the gap between the artifact and what a competent compiler engineer would produce: the same artifact appeared impressive against the prior frontier of agent capability but underwhelming against production standards. Software engineering involves non-functional requirements such as quality, reliability, and maintainability \citep{personalUtdallasEduBook} that agent-driven development may trade away. Hence, evaluations that conflate functional completion with overall quality risk overstating capability and misleading downstream decisions.

\begin{claimbox}[Document interventions]
Permit human intervention on incidental obstacles, but record precisely when, why, and how humans step in.
\end{claimbox}
Real-world tasks expose agents to obstacles incidental to the capability under test: policy refusals, CAPTCHAs, infrastructure failures. Unlike benchmarks, open-world evaluations can accommodate human intervention, which enables elicitation of upper-bound capabilities. To preserve interpretability, the degree of agent autonomy must be assessable independent of interventions. In CRUX \#1, classifying the OpenClaw daemon crash as infrastructure rather than an agent failure was a judgment call we flag for transparency; other surveyed evaluations rarely make such reasoning visible.

\begin{claimbox}[Analyze and release logs]
Treat qualitative analysis of agent logs as a first-class output, and publish the logs so external researchers can verify and extend the analysis.
\end{claimbox}
Agent logs contain substantially more information than a binary outcome and can reveal how agents decompose problems, recover from failures, explore solution spaces, and sometimes misrepresent their own progress. In CRUX \#1, log analysis surfaced both the fabricated phone number and an emergent cost optimization. Such insights are not recoverable from aggregate scores alone, and we consider systematic log analysis a defining feature of open-world evaluation. Still, runs remain hard to reproduce (recall Section~\ref{limitations-of-open-world-evaluations}) and publishing the logs does not change that. What it does enable is \emph{replicability}: external researchers can verify the analytical claims and perform analyses the original authors may not have run. To enable exactly this kind of scrutiny, we release the full CRUX \#1 logs.

\begin{claimbox}[Add real-time monitoring]
Complement post-hoc log analysis with automated real-time review, e.g., a watchdog agent that flags anomalies as they occur.
\end{claimbox}
Post-hoc log analysis is valuable but insufficient to catch all unintended agent actions. In AI Village, agents attempted to send hundreds of unsolicited emails \citep{theaidigestOrgBlogWhat}; in our own evaluation, the agent fabricated a phone number that went undetected until later review. A monitor agent reviewing the primary agent's actions in real time could surface such issues sooner than human review alone.

\begin{claimbox}[Run dry runs first]
Exercise the scaffold, evaluation criteria, and infrastructure in advance to surface implicit assumptions and scaffolding defects.
\end{claimbox}
Without this step, defects can contaminate the primary results. The two dry runs preceding CRUX \#1 identified multiple issues that were corrected before the main run.

\begin{claimbox}[Report cost]
Treat cost as a first-class quantity alongside capability.
\end{claimbox}
Agent capability on many real-world tasks continues to scale with budget. Even when a definitive upper-bound claim is not available, reporting cost-conditioned measurements allows readers to assess whether additional budget would be expected to advance task progress \citep{epochAiBlogMirrorcode}.

\newpage
\appendix

\section{Additional discussion: threats to benchmark validity}\label{app:benchmark-validity}

This appendix expands on Section~\ref{benchmarks-can-both-overestimate-and-underestimate-progress}, discussing partial remedies that have been proposed in the literature, why engineered evaluation environments remain limited as agent capabilities grow, and why fixing benchmark validity issues is structurally difficult.

\textbf{Refinements that move beyond raw accuracy.} Several recent efforts try to extract more signal from benchmark scores. \citet{arxivabs260216666} show that while agents improve rapidly on average accuracy, they improve far more slowly on metrics that capture reliability. A complementary line of work has shown that many agent-generated SWE-Bench solutions judged correct by automated test-passing would in fact be rejected by project maintainers \citep{metrOrgNotes2026}, suggesting that even outcome-oriented benchmarks miss much of what determines whether agent output is actually useful. These refinements are valuable, but they do not address what we see as the more important question for frontier evaluation: upper-bound elicitation rather than average-case measurement.

\textbf{Engineered environments and their failure modes.} As agent capabilities improve, sandboxed evaluations in domains such as coding, deep research, and customer service require increasingly engineered environments. These environments must challenge agents while also avoiding contamination and reward hacking. For example, performance on web benchmarks is affected by the frequency with which agents encounter CAPTCHAs \citep{arxivpdf251002418}, rather than by the underlying capability under test. Recent work documents agents retrieving answers online \citep{nistGovCaisiCheating}, exploiting bugs in evaluations \citep{githubComIssues465}, and producing code that passes tests but fails to meet production standards \citep{metrOrgNotes2026}. As the engineering burden grows, the real environment can become not only more valid but also more \emph{efficient} than building an elaborate sandbox; the trajectory from purely simulated retail benchmarks to Anthropic and Andon Labs' Project Vend deployments \citep{anthropicComResearchProject,anthropicComResearchProject2,andonlabsComBlogAndon} illustrates the pattern. These findings motivate a shift toward deeper qualitative evaluations that can surface failure modes and problem-solving strategies obscured by aggregate benchmark scores.

\textbf{Why fixing benchmark validity is difficult.} Qualitative log analysis \citep{aisiGovUkBlog} can surface validity concerns, but the common remedy is to release an updated benchmark, which can take months (Figure~\ref{fig:successor-benchmarks}). Open-world evaluations, by contrast, permit dry runs to identify issues before the main run, and manual intervention to resolve issues encountered during evaluation. In principle, dry runs could also be used to uncover issues with benchmarks. However, benchmarks are designed to support longitudinal comparison between models, and many validity issues only become apparent once a more capable model finds a shortcut or edge case that was not anticipated at construction time. Resolving such issues requires updating the benchmark and re-running prior models, which undermines longitudinal validity.

\textbf{Unsaturated benchmarks remain valuable.} Evaluations have always had flaws; the requirement is not perfection, but rather that they provide a useful proxy for capability progress. Several benchmarks remain unsaturated, including SciCode \citep{arxivabs240713168}, MMLU-Pro \citep{arxivabs240601574}, Humanity's Last Exam \citep{arxivabs250114249}, and SWE-Bench Pro \citep{arxivabs250916941}. Even saturated capability benchmarks can be useful for measuring efficiency and reliability, which are essential to AI diffusion. As agents become more capable, success metrics will need to become multi-faceted to capture the diversity of objectives in a given task, and benchmarks will need to incorporate bottlenecks such as human assistance and messy environments such as internet navigation. Each of these introduces its own threats to internal and external validity.

\section{CRUX \#1: extended setup and findings}\label{app:crux1-extras}

This appendix expands on Section~\ref{crux-1-can-ai-agents-autonomously-develop-and-publish-an-ios-app}, covering the agent scaffold and configuration and the dry runs.

\textbf{Agent scaffold.} We used OpenClaw \citep{openclawAi} as the agent scaffold, with Claude Opus 4.6 and adaptive thinking enabled \citep{platformClaudeComBuild}.\footnote{We made minor modifications to establish a subagent that verifies outputs and wakes the main agent every 5 minutes to check for status updates (e.g., Apple review responses). OpenClaw's default polling interval is 30 minutes; the shorter interval substantially increased the API cost of the task.} OpenClaw was a natural choice for this setting: it is configurable, integrates with the browser, and natively supports long-running tasks, all of which the iOS submission process requires. Its recent adoption also gave us a reason to evaluate it as a general-purpose scaffold, with potential comparisons to alternatives in future iterations. Aside from prompting and granting the agent deeper access to the macOS VM, we made no changes to the default OpenClaw configuration, so that the results would reflect the behavior a typical user of the scaffold could expect rather than a heavily customized setup. Part of our interest in OpenClaw stemmed from a related question: how well a modern agent would handle visual reasoning and GUI operation, which have historically been bottlenecks for computer-use agents. Prior user reports suggested that OpenClaw had partially addressed these issues, and the App Store submission process, with its screenshots, dialogs, and form-driven interactions, offered a natural setting in which to verify this. We acknowledge that OpenClaw carries security risks in real-world use; we elected to use it anyway in order to characterize the capability frontier, while noting that security may itself become a bottleneck to real-world adoption of this class of scaffold. To support all of the above, we gave the agent a macOS virtual machine with expansive permissions (sudo, screen visibility, and UI control), and logged all of its actions, reasoning traces, and screenshots.

\begin{figure}[t]
\centering
\includegraphics[width=0.6\linewidth]{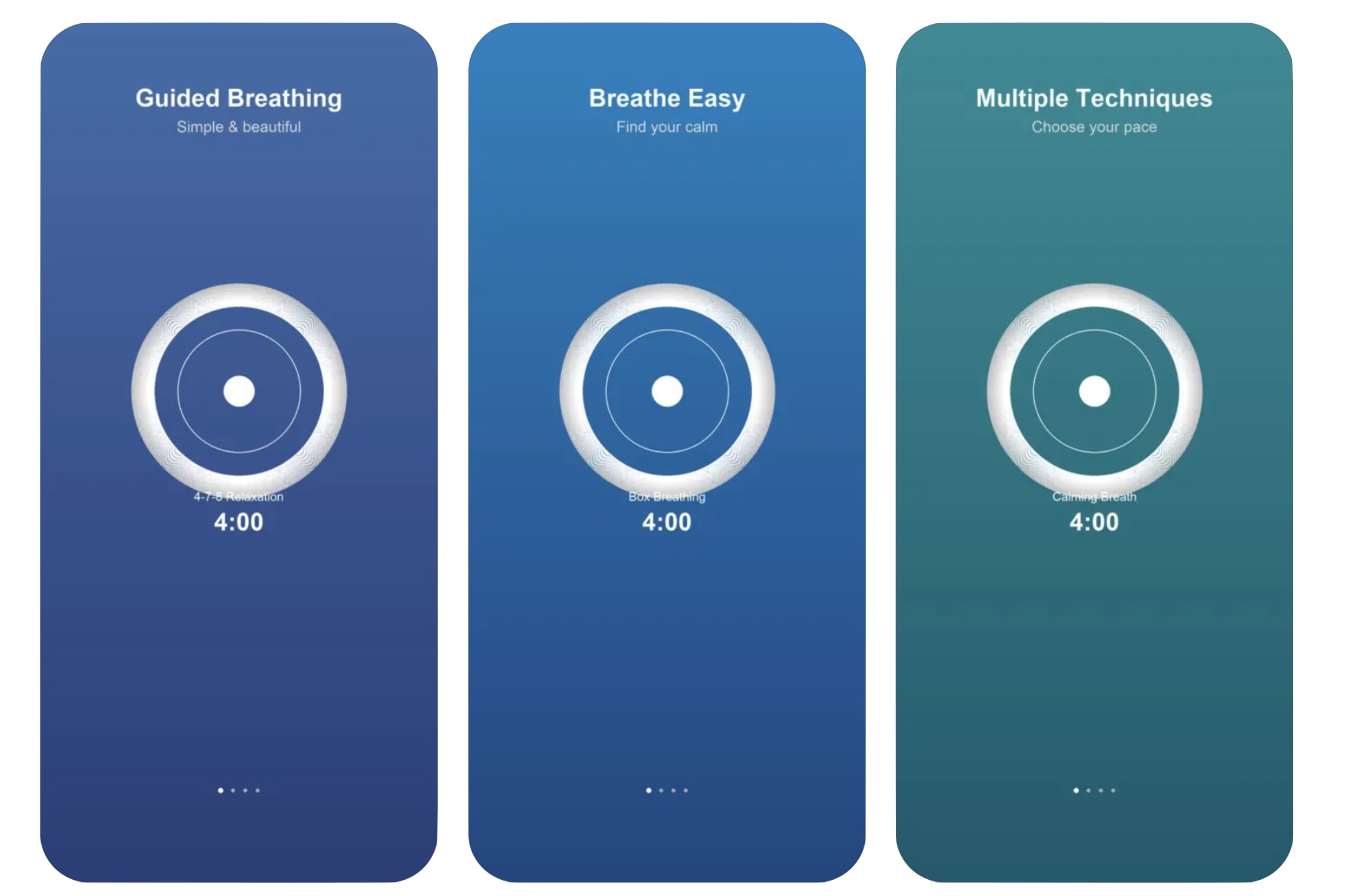}
\caption{The App Store screenshots uploaded by the agent had visible formatting errors. The application was nonetheless approved by Apple's review.}
\label{fig:screenshots}
\end{figure}

\textbf{Dry runs and setup overhead.} Prior to the main evaluation, we conducted two dry runs to verify the agent setup end-to-end and to identify scaffolding bugs that would otherwise contaminate the real run. To avoid polluting Apple's review queue with test submissions, the dry runs stopped short of actually interacting with the App Store submission or review processes. A non-trivial portion of the overall effort went into standing up the environment rather than into the task itself: establishing the OpenClaw scaffold with the necessary permissions required approximately eight person-hours of work and \$50 in API costs, covering VM configuration, the logging pipeline, and provisioning of the email, GitHub, and Apple Developer accounts. For an honest developer running a single evaluation, this setup cost is meaningful overhead. From the perspective of a prospective spammer, however, it is a one-time fixed cost, amortized across every subsequent submission, and therefore unlikely to pose a meaningful constraint on large-scale agent-driven app publication. Within the dry runs themselves, the agent relied on two primary interfaces: the command line (for code generation, build, and submission preparation) and the browser (for App Store Connect login, certificate retrieval, and form completion). When command-line operations hung awaiting GUI confirmation, for example because macOS was prompting for permission, the agent fell back on screenshots and simulated mouse clicks to approve the relevant dialogs and continue execution. This interleaving of text-based and visual interaction was itself informative, and closely resembles what we would expect from a capable human developer tackling the same workflow.

\textbf{Extended findings.} On the manual-intervention breakdown: the four non-agent interventions comprised one OpenClaw daemon crash that required a manual restart, our required pre-release approval, and Apple's synthetic-interaction block on sensitive dialogs such as two-factor authentication approval. On the credential-recovery episode: when the agent could not locate previously provided credentials, a team member suggested reusing them and resolving the associated two-factor prompt; the agent searched its own memory for the credentials but, rather than attempting a live sign-in, discovered that the App Store Connect API key was still present at the expected hidden path and used the API key to resume monitoring without ever completing an interactive sign-in. On methodological scope: the qualitative log analysis behind these findings is inherently incomplete, and other notable behaviors may remain undiscovered in the logs.

\begin{table}[t]
\centering
\caption{An incomplete survey of open-world evaluations conducted between February 2025 and March 2026.}
\label{tab:open-world-survey}
\footnotesize
\setlength{\tabcolsep}{5pt}
\begin{tabular}{@{}>{\raggedright\arraybackslash}p{2.1cm}>{\raggedright\arraybackslash}p{1.1cm}>{\raggedright\arraybackslash}p{1.5cm}>{\raggedright\arraybackslash}p{1.4cm}>{\raggedright\arraybackslash}p{3.2cm}>{\raggedright\arraybackslash}p{3.2cm}@{}}
\toprule
\textbf{Evaluation} & \textbf{Length} & \textbf{Human Role} & \textbf{Cost} & \textbf{\textcolor{capgreen}{Agent Capabilities}} & \textbf{\textcolor{limred}{Agent Limitations}} \\
\midrule
\textbf{Claude Plays Pokemon}~\citep{futurismComAdvancedAi}
& Weeks & Setup-only & Not disclosed
& \textcolor{capgreen}{Navigated menus, battled trainers, made story progress}
& \textcolor{limred}{Stuck in Mt.~Moon for $\sim$80 hours; large gap between playing and playing well} \\
\addlinespace[4pt]
\textbf{AI Village}~\citep{theaidigestOrgVillage}
& Months & Setup-only & $\sim$\$50K/yr
& \textcolor{capgreen}{Built word games; launched a Substack; late-2025 models showed meaningful improvement; sustained multi-week execution}
& \textcolor{limred}{Persistent hallucination and unproductive loops; GUI bottlenecks impeded completion of basic tasks} \\
\addlinespace[4pt]
\textbf{Project Vend}~\citep{anthropicComResearchProject,anthropicComResearchProject2}
& Weeks & Monitoring & Not disclosed
& \textcolor{capgreen}{Phase 2 achieved weekly profit; fixed Phase 1 failure modes}
& \textcolor{limred}{Social-engineering vulnerabilities: agent manipulated into giving away inventory} \\
\addlinespace[4pt]
\textbf{Cursor Browser}~\citep{cursorComBlogScaling}
& 1 week & Setup-only & $\sim$3B tokens (\$10K--50K)
& \textcolor{capgreen}{Functional Rust rendering engine supporting HTML, CSS, layout, and paint on real sites}
& \textcolor{limred}{Far from production quality; flat swarm collapsed to 2--3 effective agents before a hierarchical reorganization} \\
\addlinespace[4pt]
\textbf{C Compiler}~\citep{anthropicComEngineeringBuilding}
& 2 weeks & Monitoring & $\sim$\$20K
& \textcolor{capgreen}{99\% GCC torture-test pass rate; compiled the Linux kernel, PostgreSQL, Redis, FFmpeg, and Doom}
& \textcolor{limred}{Less efficient than GCC with optimizations disabled; ceiling around $\sim$100K lines of code at which bug fixes broke existing functionality} \\
\addlinespace[4pt]
\textbf{Epoch knowledge-work tasks}~\citep{epochAiGradientUpdates}
& Hours & Setup-only & \$20--200/mo
& \textcolor{capgreen}{Partial success on Substack porting and web-interface replication}
& \textcolor{limred}{Failed at basic GUI operations; hallucinated data; visual computer-use horizons 40--100$\times$ shorter than text-based} \\
\addlinespace[4pt]
\textbf{Codex Design Tool}~\citep{developersOpenaiComBlog}
& $\sim$25 hrs & Setup-only & $\sim$\$200
& \textcolor{capgreen}{Long-horizon coherence via milestone-based planning and verification}
& \textcolor{limred}{Reported only by authors; no independent evaluation or live demonstration} \\
\addlinespace[4pt]
\textbf{vinext}~\citep{blogCloudflareComVinext}
& $\sim$1 week & Collaborative & $\sim$\$1{,}100
& \textcolor{capgreen}{94\% Next.js API coverage; 4.4$\times$ faster builds; 57\% smaller bundles}
& \textcolor{limred}{Target was well-specified with existing test suites; Vite and its RSC plugin provided much of the functionality} \\
\addlinespace[4pt]
\textbf{Training a Computer}~\citep{xDimitrisPapail2028669695344148946}
& Hours--days & Mixed & \$20--200/mo
& \textcolor{capgreen}{Human-guided Claude Code generalized to multi-step computations absent from training}
& \textcolor{limred}{Both agents reward-hacked in the fully autonomous condition} \\
\addlinespace[4pt]
\textbf{Nanochat Autoresearch}~\citep{xkarpathy2031135152349524125}
& Days & Collaborative & $<$\$100
& \textcolor{capgreen}{$\sim$100 experiments overnight; reported 11--19\% improvements on ``Time to GPT-2''}
& \textcolor{limred}{Absolute improvements are small; the agent lacks a mechanism for reasoning about why a change succeeded} \\
\bottomrule
\end{tabular}
\end{table}

\section{Survey of open-world evaluations: details}\label{app:survey}

This appendix expands on Section~\ref{an-incomplete-survey-of-open-world-evaluations} by describing ten representative open-world evaluations conducted between February 2025 and March 2026, and provides a side-by-side comparison of their reported capabilities, limitations, and costs in Table~\ref{tab:open-world-survey}.

\begin{enumerate}[leftmargin=16pt]
\def\labelenumi{\arabic{enumi}.}
\item \textbf{Anthropic, Claude Plays Pokemon} \citep{futurismComAdvancedAi} (February 2025). Anthropic ran a Twitch livestream in which Claude 3.7 Sonnet played Pokemon Red. Although not a deployment, the experiment was an early instance of situating an agent in a relatively open environment compared to typical benchmarks. It illustrated both progress in AI computer use under minimal scaffolding and the limitations of early-2025 agents: the agent remained stuck on a single level for approximately 80 hours \citep{futurismComAdvancedAi}.\footnote{For context, most children are able to beat the entire game in around 25 hours \citep{howlongtobeatComGame7169}.}
\item \textbf{AI Digest, AI Village} \citep{theaidigestOrgVillage} (April 2025--present). Multiple agents are given individual computer environments and a shared group chat, and tasked with open-ended real-world goals such as charity fundraising, organizing in-person events, and building a Substack presence. The project has documented persistent failure modes such as hallucination, miscalibration, and unproductive loops, alongside notable improvements from late-2025 agents \citep{theaidigestOrgBlogWhat}.
\item \textbf{Anthropic/Andon Labs, Project Vend} \citep{anthropicComResearchProject} (June 2025--present). Anthropic and Andon Labs deployed a Claude 3.7 Sonnet-based agent (``Claudius'') to operate a small automated store in Anthropic's office, managing inventory, pricing, and customer interactions over several weeks, and surfacing failure modes around manipulation, prioritization, and real-world decision-making. A second phase \citep{anthropicComResearchProject2} expanded the experiment to multiple locations with newer models and included a red-teaming exercise by Wall Street Journal staff. The original phase incurred substantial losses from poor planning, hallucination, and excessive discounting; the follow-up was more profitable, though WSJ staff were still able to jailbreak the agent into giving away inventory \citep{futurismComFutureSociety}. Andon Labs has since initiated a third phase in which a Claude-based agent (``Luna'') has been given a three-year lease on a brick-and-mortar store and tasked with operating it profitably, including hiring employees and designing the brand \citep{andonlabsComBlogAndon}.
\item \textbf{Cursor, browser experiment} \citep{cursorComBlogScaling} (January 2026). Wilson Lin at Cursor coordinated hundreds of GPT-5.2 agents to build a web browser from scratch over one week. The resulting browser (``FastRender'') comprised over a million lines of Rust, including a from-scratch rendering engine. It could render simple websites but was far from production-ready. The experiment is notable for its exploration of hierarchical multi-agent coordination at scale and for characterizing the failure modes that emerge over multi-day agent runs.
\item \textbf{Carlini, C compiler} \citep{anthropicComEngineeringBuilding} (February 2026). Nicholas Carlini used Claude to build a C compiler from scratch at a cost of approximately \$20{,}000 in API usage. The agent produced a working compiler capable of compiling the Linux kernel and passed a large fraction of standard test suites. The experiment surfaced both strengths (systematic code generation, test-driven iteration) and weaknesses (complex optimization passes, debugging subtle specification violations).
\item \textbf{Ho, knowledge-work tasks at Epoch} \citep{epochAiGradientUpdates} (February 2026). Anson Ho had Claude Code and ChatGPT Atlas attempt three knowledge-work tasks at Epoch: replicating an interactive web interface for a 40-parameter economic model, writing a 2025 AI-progress article in Epoch's style, and porting an article from Google Docs to Substack and Epoch's website. Formatting bottlenecks and hallucinations emerged as persistent limitations.
\item \textbf{Choi, GPT-5.3 Codex design tool} \citep{developersOpenaiComBlog} (February 2026). Derrickk Choi of OpenAI ran GPT-5.3 Codex autonomously for 25 hours to generate 35{,}000 lines of code for a ``design tool.'' The associated report describes planning, memory, and verification behavior but does not substantively analyze the capabilities and limitations of the finished product.
\item \textbf{Faulkner, Next.js reimplementation} \citep{blogCloudflareComVinext} (February 2026). A Cloudflare engineer used Claude with OpenCode to release \emph{vinext}, a reimplementation of the Next.js frontend framework atop Vite rather than React. The associated writeup reported 94\% coverage of Next.js for approximately \$1{,}100 in API costs, but subsequent community analysis \citep{newsYcombinatorComItem} identified security limitations and limited generalizability, noting that much of the heavy lifting was provided by the existing testing infrastructure of Vite and Next.js.
\item \textbf{Papailiopoulos et al., training a computer} \citep{xDimitrisPapail2028669695344148946} (March 2026). The authors tested whether Claude Code and OpenAI Codex could train a transformer to function as a general-purpose computer. In the fully autonomous condition, both agents failed and produced reward-hacked solutions; in a human-guided condition, Claude Code succeeded and displayed meaningful generalization, including to multi-step computations absent from its training.
\item \textbf{Karpathy, Nanochat autoresearch} \citep{xkarpathy2031135152349524125} (March 2026). Karpathy built a simple automation pipeline atop the open-source nanochat project (for GPT-2-level LLM training), giving an agent full autonomy to adjust architecture, hyperparameters, optimizers, and batch sizes in 5-minute increments. In a follow-up, Karpathy reported that the agent reduced the ``Time to GPT-2'' metric (measured on $8\times$H100 GPUs) by 11\% over two days.
\end{enumerate}

\end{document}